\documentclass{article}
\usepackage{spconf,amsmath,graphicx}
\usepackage{booktabs}
\usepackage{multirow}
\usepackage{eso-pic}
\usepackage{microtype}
\usepackage{pifont}

\usepackage[table]{xcolor}

\definecolor{best}{HTML}{C6EFCE}  
\definecolor{second}{HTML}{FFEB9C} 

\usepackage{pgf}
\usepackage{xcolor}
\usepackage{xspace} 

\definecolor{low}{HTML}{F8696B}   
\definecolor{high}{HTML}{63BE7B}  

\newcommand{\improvement}[1]{%
    \pgfmathparse{#1}%
    \let\val\pgfmathresult%
    \ifdim \val pt < 0 pt%
        \pgfmathparse{int(min(100, -1*\val/12.2*100))}%
        \edef\temp{\noexpand\cellcolor{low!\pgfmathresult!white}}%
        \temp #1\%%
    \else%
        \pgfmathparse{int(min(100, \val/28.5*100))}%
        \edef\temp{\noexpand\cellcolor{high!\pgfmathresult!white}}%
        \temp #1\%%
    \fi%
}

\newcommand{\bg}[1]{\cellcolor{best}#1}   
\newcommand{\sy}[1]{\cellcolor{second}#1} 
\renewcommand{\bg}[1]{#1} 
\renewcommand{\sy}[1]{#1}

\newcommand{\methodname}{MDE-VIO\xspace}
\usepackage{url}

\title{\methodname: ENHANCING VISUAL-INERTIAL ODOMETRY USING LEARNED DEPTH PRIORS}
\name{Arda Alnıak$^{1,3}$ \quad Sinan Kalkan$^{2,3}$ \quad M. Mert Ankaralı$^{1,3}$ \quad Afşar Saranlı$^{1,3}$ \quad A. Aydın Alatan$^{1,3}$ }

\address{$^1$Dept. of Electrical \& Electronics Engineering, $^2$Dept. of Computer Engineering \\ 
$^3$Center for Image Analysis (OGAM), METU, Ankara, Türkiye}

\begin{document}

\maketitle

\begin{abstract}
Traditional monocular Visual-Inertial Odometry (VIO) systems struggle in low-texture environments where sparse visual features are insufficient for accurate pose estimation. To address this, dense Monocular Depth Estimation (MDE) has been widely explored as a complementary information source. While recent Vision Transformer (ViT) based complex foundational models offer dense, geometrically consistent depth, their computational demands typically preclude them from real-time edge deployment. Our work bridges this gap by integrating learned depth priors directly into the VINS-Mono optimization backend. We propose a novel framework that enforces affine-invariant depth consistency and pairwise ordinal constraints, explicitly filtering unstable artifacts via variance-based gating. This approach strictly adheres to the computational limits of edge devices while robustly recovering metric scale. Extensive experiments on the TartanGround and M3ED datasets demonstrate that our method prevents divergence in challenging scenarios and delivers significant accuracy gains, reducing Absolute Trajectory Error (ATE) by up to 28.3\%. Code will be made available.
\end{abstract}

\section{Introduction}
\label{sec:introduction}

\begin{figure}
    \centering
    \includegraphics[width=0.8\linewidth]{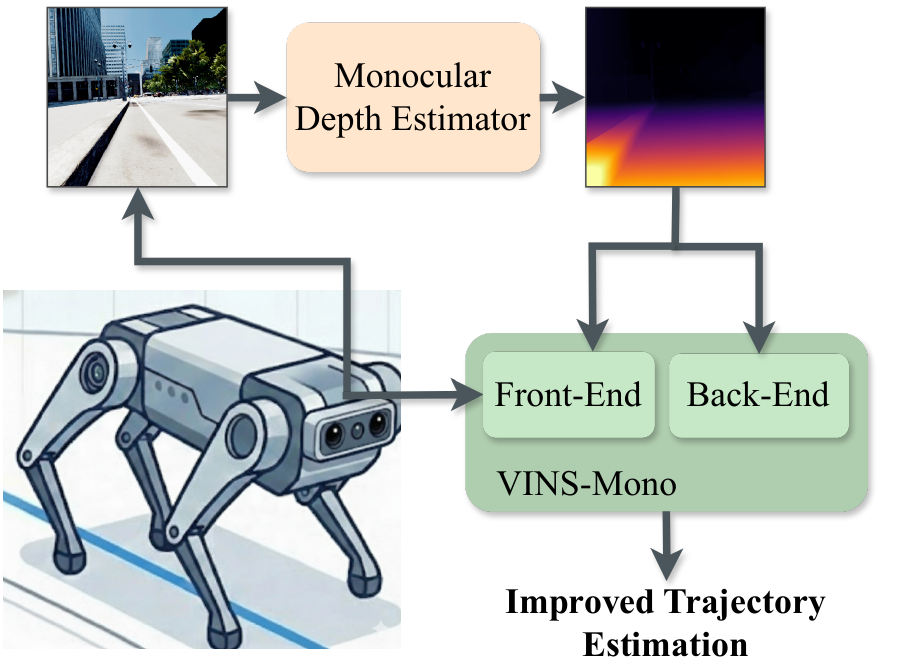}
    \caption{\textbf{High-level overview of the proposed \methodname framework.} The system augments the standard VINS-Mono by processing monocular images through a Monocular Depth Estimator. These learned depth priors are integrated into both the Front-End and Back-End of the estimator to achieve robust, metric-scaled trajectory estimation on robotic platforms.}
    \label{fig:teaser}
\end{figure}


Odometry, the process of estimating sequential changes in pose from sensor data, is fundamental for precise localization in autonomous systems. Among various sensor options for odometry, monocular cameras are ubiquitous due to their low weight and power requirements. However, recovering absolute metric scale from a single camera is an ill-posed problem \cite{slam-handbook}. Hence, Visual-Inertial Odometry (VIO) systems fuse visual data with Inertial Measurement Unit (IMU) readings for obtaining a metric-scaled trajectory \cite{slam-handbook}. However, traditional optimization-based VIO systems often suffer from degraded accuracy in low-texture environments where visual features are sparse \cite{peng2022rwtslamrobustvisualslam}.

Recent advances in Transformer-based architectures have revolutionized computer vision. Foundational models, such as Vision Transformers (ViT) \cite{vit} and the DINO family \cite{dinov2, dinov3}, have enabled highly accurate, dense Monocular Depth Estimation (MDE) \cite{daac, dav2, dav3, depthpro, vda, metric3dv2}.
While these architectures have paved the way for end-to-end learning-based odometry \cite{droidslam, dpvo}, such ``black-box'' systems are often computationally prohibitive for resource-constrained edge devices (e.g., DROID-SLAM demands dual RTX 3090 GPUs to achieve 20 FPS \cite{droidslam}). 

In this work, we bridge this gap by integrating lightweight, learning-based depth priors into a traditional optimization-based VIO framework. Unlike many end-to-end approaches that replace the entire pipeline, our method enhances the front-end inputs and backend optimization of VINS-Mono \cite{vinsmono}, ensuring suitability for embedded applications. Our primary contributions are as follows:

\noindent\textbf{(1)} We integrate a state-of-the-art MDE into an optimization-based VIO system, specifically tailored for the computational constraints of edge devices, e.g., NVIDIA Jetson AGX Orin.

\noindent\textbf{(2)} We propose two novel depth integration methods: \textbf{D}epth-\textbf{I}njected \textbf{F}eature \textbf{T}racking (\textbf{DIFT}), which embeds geometric priors into the visual channel to enhance gradient detection in texture-poor regions, and \textbf{Or}dinal \textbf{C}onstraints (\textbf{OrC}) to enforce relative depth ordering consistency in the factor graph. We show that diverse integration strategies can unlock distinct performance gains, demonstrating that front-end injection provides a novel solution for texture-poor image robustness while backend constraints ensure metric precision, thus establishing new directions for effective MDE-VIO fusion on edge devices.
    
\noindent\textbf{(3)} We analyze the impact of flicker in zero-shot depth models versus video-based models on VIO integration. We introduce an uncertainty-based dynamic weighting mechanism to filter unstable depth artifacts, ensuring that only reliable geometric priors influence the optimization for more robust use.


\section{RELATED WORK}
\label{sec:format}

\textbf{Monocular Depth Estimation (MDE).} MDE has evolved from early convolutional approaches~\cite{eigen} to ViT-based architectures~\cite{dav2, dav3, daac, moge, vda}. Affine-invariant models~\cite{eigen, dav2, dav3, daac, moge, vda} employ invariant losses, where predicted inverse depth $\hat{d}$ relates to metric depth $d^{*}$ via $d^{*} = s\hat{d} + t$. In contrast, metric MDE infers absolute scale; while early methods required known intrinsics~\cite{metric3d, unidepth}, recent models estimate them during inference~\cite{metric3dv2, depthpro, unidepthv2}. Moreover, high-generalization affine models are increasingly fine-tuned for metric accuracy~\cite{dav2, dav3, vda}.

\noindent\textbf{Optimization-based VIO}. VIO fuses visual and inertial data for precise trajectory estimation. Unlike computationally heavy direct methods, indirect approaches balance accuracy and efficiency. Optimization-based indirect systems~\cite{vinsmono} estimate the optimal state using a factor graph to minimize the joint non-linear cost function:
\begin{equation}\footnotesize
    J(\mathbf{x}) = \sum \| \mathbf{r}_{\mathcal{C}}(\mathbf{x}) \|^2_{\mathbf{w}} + \sum \| \mathbf{r}_{\mathcal{I}}(\mathbf{x}) \|^2_{\mathbf{\Sigma}_{\mathcal{I}}} + \dots
\end{equation}
where $\mathbf{r}_{\mathcal{C}}$ penalizes the misalignment between 3D landmarks and their 2D projections; $\mathbf{r}_{\mathcal{I}}$ constrains relative motion using preintegrated IMU measurements. These terms are weighted by the information matrices $\mathbf{w}$ and $\mathbf{\Sigma}_{\mathcal{I}}^{-1}$, respectively.

\begin{table}[ht] 
\centering
\caption{A Summary of MDE-aided optimization-based VIO systems: (*) Requires prior knowledge of the test domain to yield accurate metric depth; resulting in limited generalization.}
\label{tab:related_work}
\footnotesize\setlength{\tabcolsep}{2pt}
\begin{tabular}{cccl}
\hline
\textbf{Work} & \textbf{Metric} & \textbf{   Edge Deployable   } & \textbf{Other Limitation} \\
\hline
\cite{pseudorgb} & \ding{51}* & \ding{55} & High compute (offline training) \\
\cite{onlinetrain} & \ding{51}* & \ding{55} & High compute (online training) \\
\cite{mdvo} & \ding{51}* & \ding{55} & High compute (dual complex MDEs) \\
\cite{dfvo} & \ding{51}* & \ding{55} & High compute (heavy architecture) \\
\cite{dptvo} & \ding{51}* & \ding{51} & Limited generalization \\
\cite{triangulation} & \ding{51}* & \ding{55} & High compute (heavy architecture) \\
\cite{carlos} & \ding{51}* & \ding{55} & High compute (online training) \\
\cite{ordinal} & \ding{55} & \ding{55} & High compute (heavy depth model) \\
\cite{tightlycoupled} & \ding{51} & \ding{55} & High compute (heavy architecture) \\
\hline
\textbf{Ours} & \ding{51} & \ding{51} & \textbf{None (Real-time backend)} \\
\hline
\end{tabular}
\end{table}

\noindent\textbf{MDE-aided VIO Systems.} Approaches for integrating MDE into VIO include test-time training for domain adaptation \cite{pseudorgb, onlinetrain}, ordinal-based outlier rejection \cite{ordinal}, and fusion with multiple networks \cite{mdvo} or optical flow \cite{dfvo}. Others recover affine parameters by aligning dense predictions to sparse landmarks \cite{dptvo, triangulation} or leverage multi-view constraints \cite{tightlycoupled}. However, these methods largely ignore edge device constraints, failing to provide real-time, metric-accurate estimation on resource-limited hardware.

\begin{figure*}[t]
    \centering
    \includegraphics[width=0.9\textwidth]{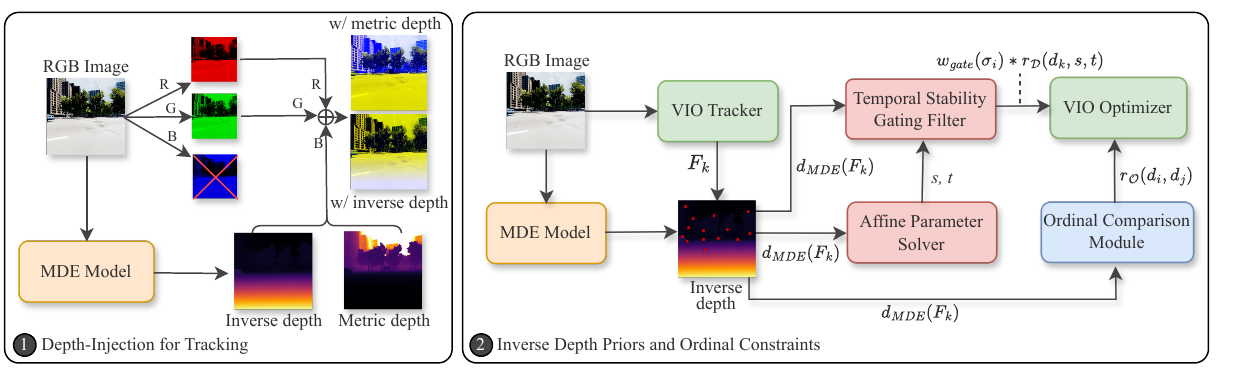}
    \caption{\methodname: The proposed VIO enhancement framework. (\textbf{Left}) The DIFT module replaces the Blue channel of the RGB input with normalized MDE predictions to enhance KLT tracking in low-texture regions. (\textbf{Right}) The proposed approach aligns learned depth priors $d_{MDE}$ for the set of tracked features $\mathcal{F}_k$ using affine parameters $(s, t)$, filters unstable estimates via a variance-based gate $w_{gate}(\sigma_i^2)$, and integrates both unary depth factors $r_{\mathcal{D}}$ and pairwise OrC $r(d_i, d_j)$ into the VIO Optimizer.}
    \label{fig:two_column}
\end{figure*}

\section{PROPOSED METHOD}
\label{sec:method}
\subsection{System Overview and Design Choices}

As summarized in Fig. \ref{fig:two_column}, \methodname augments VINS-Mono by embedding depth priors into the frontend via Depth-Injected Feature Tracking (DIFT) and constraining the backend through variance-gated affine ($s, t$) and pairwise ordinal ($r_{\mathcal{O}}$) residuals.

For \methodname, we selected VINS-Mono \cite{vinsmono} for being a popular, high-performance and robust VIO algorithm. Another important factor for this decision was VINS-Mono's superiority in our experiments at initialization compared to other approaches, e.g., ORB-SLAM3 \cite{orbslam}.


As we aim to work on edge devices, it is not feasible to employ any diffusion-based models, as well as any complex MDE models with more parameters. Conversely, optimization demands highly accurate information, which requires a rigorous method selection considering speed and accuracy trade-off. Our preliminary experiments revealed that direct monocular metric depth estimation is insufficient for enhancing VIO; the raw depth outputs often exhibit scale inconsistencies relative to the IMU, degrading the optimization results. Since affine alignment is inevitable, we prioritize the generalization 
capabilities of affine-invariant models: DepthAnythingAC \cite{daac} and VideoDepthAnything \cite{vda}. 

Consequently, our backend integration targets the refinement of the estimator's core state variables. Since VINS-Mono parameterizes 3D landmarks using \textit{inverse depth}, we leverage the MDE geometric priors to explicitly constrain these inverse depth states within the backend optimization.

\subsection{Depth-Injected Feature Tracking (DIFT)}
Processing dense depth maps as a separate modality incurs significant overhead on edge devices. To leverage geometric priors without increasing input dimensionality, we propose a channel-embedding strategy. We apply $2\%$-$98\%$ percentile clamping to the depth predictions (metric or inverse) for robust normalization against outliers, convert the result to 8-bit intensity, and use it as the Blue channel of the RGB input (Fig. \ref{fig:two_column} (left)): $RGB \rightarrow RGD$. Since standard grayscale conversion weights the Blue channel least ($0.114$), this operation effectively superimposes geometric structure onto the visual texture with minimal photometric disruption via $I_{track} = 0.299R + 0.587G + 0.114D_{norm}$. This composite representation enhances gradient magnitudes in texture-less regions, thereby improving KLT tracking within VINS-Mono in geometrically distinct but visually flat areas.

\subsection{Uncertainty-guided Dynamic Adaptation}
We utilize DepthAnything at Any Condition (DAAC) \cite{daac} and VideoDepthAnything (VDA) \cite{vda} as MDE models. DAAC promises strong generalization, yet suffers from inherent inter-frame flickering that limits its usability. VDA is a video depth estimation model providing temporal stability. We employ VDA's streaming feature\footnote{The authors \cite{vda} argue that this reduces the stability of consecutive depth estimations. Aligning with that argument, our tests revealed lesser but distinguishable inter-frame flickering.}. To differentiate between reliable estimations and artifacts, we implement a temporal consistency check for each feature track along its trajectory. We compute the sample variance $\sigma^2_i$ of the inverse depth predictions over a sliding window as a measure for network uncertainty. This variance modulates the weight of the depth prior in the factor graph of optimization.

In standard operation, if a feature $f_i$'s depth hypothesis variance $\sigma_i^2$ exceeds a critical outlier threshold $\sigma_{thresh}^2$, the prior is deemed unreliable and excluded from the optimization graph entirely.
We seek to maximize information usage even from noisy features, thus features passing hard rejection are subject to an uncertainty-based weighting function. We model the confidence weight $w_i$ by using a sharp Gaussian kernel that rapidly decays as variance increases:
\begin{equation}\footnotesize
w_{gate}(\sigma_i) = \exp\left(-\gamma \cdot \sigma_i^2\right),
\end{equation}
where $\gamma$ is a scaling factor (e.g., $\gamma = 10^4$).  In the global optimization (Eq. 7), this weight modulates the stiffness of the depth consistency residual $r_{\mathcal{D}}$. This ensures that highly unstable features possess low information values, while consistent features exert strong constraints on the state estimation.
\begin{table*}[t]
\centering
\caption{
    Quantitative results TartanGround Dataset (Downtown): ATE RMSE in meters, median of 10 runs.  
    Weights: Depth=300, Ordinal=10. 
    \textbf{Bold}/\underline{underline}: best/second best.
}
\label{tab:tartanground}\footnotesize
\setlength{\tabcolsep}{3pt} 
\begin{tabular}{llcccccccccc}
\toprule
\textbf{MDE} & \textbf{Integration} & \textbf{P2001} & \textbf{P2002} & \textbf{P2003} & \textbf{P2004} & \textbf{P2005} & \textbf{P2006} & \textbf{P2007} & \textbf{P2008} & \textbf{Avg} & \textbf{Improvement} \\
\midrule
- & VINS-Mono (Baseline) & 0.80 & 0.57 & 1.38 & 0.35& 0.38 & 0.88 & 0.24 & 0.81 & 0.676 & - \\
\midrule 
\multirow{6}{*}{DAAC} 
  & MDI & 0.68 & 0.53 &  1.24 & 0.35 & 0.37 & 0.75 & 0.26 & 0.83 & 0.626 & \improvement{+7.4}  \\
  
  & Depth Residuals + MDI & 0.43 & 0.53 & 0.99 & \textbf{0.31} & 0.24 & 0.56 & 0.26 & 0.82 & 0.518 & \improvement{+23.4}  \\
  
 & OrC + MDI & 0.57 & 0.49 & 1.22 & 0.33 & 0.36 & 0.75 & 0.26 & 0.77 & 0.594 & \improvement{+12.2} \\
 
   & Depth Residuals + OrC + MDI & 0.53 & 0.47 & 1.07 & \underline{0.32} & 0.26 & 0.50 & 0.26 & 0.87 & 0.535 & \improvement{+20.9} \\
   
 & Inverse DIFT & 0.83 & 1.06 & 1.04 & 0.43 & 0.53 & 0.82 & 0.55 & \sy{\underline{0.70}} & 0.751 & \improvement{-10.2} \\
 
 & Metric DIFT & 0.69 & 0.71 & 1.44 & 0.77 & 0.46 & 0.55 & 0.24 & 1.21 & 0.759 & \improvement{-12.2} \\
 
\midrule
\multirow{6}{*}{VDA} 
  & MDI & 0.57 & 0.51 & 1.19 & 0.35 & 0.35 & 0.76 & 0.23 & 0.81 & 0.596 & \improvement{+11.9}\\
   & Depth Residuals + MDI & \textbf{\bg{0.27}} & 0.49 & \sy{\underline{0.97}} & 0.38 & \bg{\textbf{0.23}} & \underline{\sy{0.43}} & 0.25 & 0.80 & \bg{\textbf{0.478}} & \improvement{+28.3}\\
 & OrC + MDI & 0.43 & 0.48 & 1.01 & 0.38 & 0.35 & 1.01 & 0.23 & 0.75 & 0.580 & \improvement{+7.9} \\
  & Depth Residuals + OrC + MDI & 0.34 & \sy{\underline{0.46}} & 1.04 & \sy{\underline{0.34}} & \sy{\underline{0.24}} & 0.55 & \sy{\underline{0.20}} & 0.78 & 0.508 & \improvement{+24.6} \\
 & Inverse DIFT & 0.68 & 0.71 & \bg{\textbf{0.68}} & 0.42 & 0.29 & 0.61 & \bg{\textbf{0.09}} & \bg{\textbf{0.54}} & \underline{0.502} & \improvement{+24.9} \\
 & Metric DIFT & 0.59 & \bg{\textbf{0.44}} & 1.42 & 0.44 & 0.46 & \bg{\textbf{0.35}} & 0.24 & 1.27 & 0.651 & \improvement{+3.7} \\
\bottomrule
\end{tabular}
\end{table*}
\subsection{MDE-assisted Depth Initialization (MDI)}
Standard monocular VIO often requires sufficient parallax to triangulate new features, often disabling them in optimization for a few frame after their occurrence, or not considering them at all. This limits the contribution of these features, while headstarting them is possible using MDE information. To accelerate this process, we implement a probabilistic rescue mechanism that initializes the inverse depth value faster than triangulation method. Instead of waiting for triangulation geometric convergence, our method considers the hypothetical reprojection error of these features between two frames, and initialize each feature with an error under a  threshold, ${\tau}$:
\begin{equation}\footnotesize
    \left\| \pi \left( \mathbf{T}_{ji} \left[ \hat{d}^{-1}_{\text{init}} \pi^{-1}(\mathbf{u}_i) \right] \right) - \mathbf{u}_j \right\|_2 < \tau.
\end{equation}
To handle depth uncertainty at boundaries, we select the maximum inverse depth within a $5 \times 5$ patch, which favors features belonging to the foreground. 

\subsection{Affine-Invariant Depth Residuals}
\label{ssec:affine_integration}
To effectively utilize affine-invariant network predictions, one must robustly align these predictions to the metric scale of the VIO system. We assume the global scale $s$ and shift $t$ evolve smoothly over time and solve for them using the set of correspondences $\mathcal{C} = \{(d_i, \hat{d}_i) | i \in \mathcal{F}\}$ between the triangulated metric VIO depths $d_i$ and the sampled MDE network predictions $\hat{d}_i$, where $\mathcal{F}$ denotes the set of tracked features in the sliding window.

Given that MDE networks often exhibit artifacts, we employ a two-stage estimation pipeline to reject outliers. First, we apply a Random Sample Consensus (RANSAC) scheme to identify a robust inlier set $\mathcal{I} \subset \mathcal{C}$, strictly enforcing physical consistency (i.e., $s > 0$). Second, we compute the optimal measurement pair $(s_{meas}, t_{meas})$ via a closed-form least squares refinement on the inlier set. To prevent frame-to-frame affine parameter flicker and ensure global alignment stability, we update the system's affine state using an exponential moving average (EMA) with a smoothing factor $\alpha$:
\begin{equation}\footnotesize
\begin{split}
    s_{k} &= (1-\alpha)s_{k-1} + \alpha s_{meas}, \\
    t_{k} &= (1-\alpha)t_{k-1} + \alpha t_{meas}.
\end{split}
\end{equation}

This temporal smoothing ensures that the aligned depth provides a consistent geometric constraint to the backend. Finally, we incorporate this information into the VINS-Mono sliding-window optimization by defining a depth consistency residual $r_{\mathcal{D}}$. $r_{\mathcal{D}}$ minimizes the discrepancy between the inverse depth state variable $d_k$ and the aligned network prediction:
\begin{equation}\footnotesize
    r_{\mathcal{D}}(d_{k}, s, t) = d_{k} - (s \cdot d_{MDE} + t).
\end{equation}
By integrating $r_{\mathcal{D}}$ as a unary factor, we constrain the sparse map optimization with dense geometric priors while maintaining the system's real-time feasibility.

\subsection{Pair-wise Ordinal Depth Constraints (OrC)}
Lightweight MDE predictions are prone to noise but preserve reliable relative ordering. We exploit this via pair-wise ordinal constraints, penalizing the optimizer when the estimated depth order contradicts the network's local geometric consistency.

Enforcing pairwise ordinal depth constraints between all feature combinations is computationally unfeasible ($\mathcal{O}(N^2)$). Instead, we employ a spatio-temporal filter to select a sparse set of geometrically informative constraints. Let $\mathcal{F}_k$ denote the set of features observed in the current frame $k$.
For every pair of features $f_i, f_j \in \mathcal{F}_k$, we generate a candidate constraint only if they satisfy strict spatial and geometric criteria:
 
\noindent\textbf{(i) Locality}: The Euclidean distance between their pixel coordinates must be less than a threshold $\tau_{\text{dist}}$. This excludes distant regions as they are likely unrelated.

\noindent\textbf{(ii) Significant Margin}: To ensure ordinal reliability against prediction noise, we require a minimum disparity $| \hat{d}_i - \hat{d}_j | > \eta \cdot \max(\hat{d}_i, \hat{d}_j)$, where $\eta$ represents a safety margin (e.g., $0.05$).

For temporal stability, we verify the ordinal relationship across time. Let $\mathcal{K}_{ij}$ be the set of previous frames where both $f_i$ and $f_j$ were observed. The pair is accepted if and only if the sign of their relative depth remains consistent for all $t \in \mathcal{K}_{ij}$.
A single sign reversal indicates independent motion or depth map jitter, triggering the immediate rejection of the pair.

Accepted pairs are integrated into the factor graph using a Soft Hinge Loss to penalize when the VIO state violates the ordinal inequality $d_i > d_j$ established by the network:
\begin{equation}\footnotesize
    r_{\mathcal{O}}(d_i, d_j) = \max\left(0, (d_j + \epsilon) - d_i\right),
    \label{eq:ordinal_residual}
\end{equation}
where $d_i, d_j$ represent the inverse depths in the state vector. The margin $\epsilon$ is adaptive, scaled by the depth magnitude ($\epsilon = 0.02 \cdot d_i$) to account for imperfection of MDEs.

\subsection{Adjusted Optimization Cost Function}
The final state estimation is obtained by solving an optimization problem incorporating our depth-derived constraints:
\begin{equation} \footnotesize
    J = 
    \sum \| \mathbf{r}_{\mathcal{C}} \|^2_{\mathbf{w}} + 
    \sum \| \mathbf{r}_{\mathcal{I}} \|^2_{\mathbf{\Sigma}_{\mathcal{I}}} + 
    \sum \| r_{\mathcal{D}} \|^2_{w_{g}} + 
    \sum \| r_{\mathcal{O}} \|^2_{w_{\mathcal{O}}},
    \label{eq:total_optimization}
\end{equation}
where $\mathbf{r}_{\mathcal{C}}$ and $\mathbf{r}_{\mathcal{I}}$ represent the standard visual reprojection and IMU preintegration residuals, weighted by the information matrices $\mathbf{w}$ and $\mathbf{\Sigma}_{\mathcal{I}}^{-1}$, respectively. The proposed depth consistency residual $r_{\mathcal{D}}$ minimizes the alignment error between the state inverse depth $d_i$ and the learned prior, scaled by the dynamic variance-based weight $w_{gate}(\sigma_{i})$. Finally, $r_{\mathcal{O}}$ enforces pairwise relative depth order between features $d_i, d_j$, weighted by a constant factor $w_{\mathcal{O}}$.

\section{EXPERIMENTAL RESULTS}
\label{sec:experiment}
\begin{table*}[t]
\centering
\caption{
    Quantitative results on M3ED: ATE RMSE in meters, median of 10 runs. 
    Columns use aliases (e.g., `Penno' for `penno\_short\_loop'). 
    Weights: Depth=100, Ordinal=10.  
    \textbf{Bold}/\underline{underline}: best/second best. 
    \textbf{X} denotes divergence. 
    \textbf{Note:} VDA-based methods successfully track `Stairs' where others diverge; summary improvement metrics exclude this sequence. VDA-based DIFT approach is not real-time feasible in this dataset (25 FPS), thus we do not present its results.
}
\label{tab:m3ed}
\setlength{\tabcolsep}{4pt}\footnotesize
\begin{tabular}{llccccccccc}
\toprule
\multirow{2}{*}{\textbf{MDE}} & \multirow{2}{*}{\textbf{Method}} & \multicolumn{2}{c}{\textbf{Forest}} & \multicolumn{3}{c}{\textbf{Indoor}} & \multicolumn{2}{c}{\textbf{Outdoor}} & \multicolumn{2}{c}{\textbf{Summary}} \\
\cmidrule(lr){3-4} \cmidrule(lr){5-7} \cmidrule(lr){8-9} \cmidrule(lr){10-11}

 & & \textbf{Hard} & \textbf{Road} & \textbf{Stairs} & \textbf{Build} & \textbf{Stairwell} & \textbf{Penno} & \textbf{Skate} & \textbf{Avg} & \textbf{Improvement} \\
\midrule

- & VINS-Mono (Baseline) & 0.57 & 0.73 & X & 0.31 & 0.53 & 0.60 & 0.15 & 0.482 & - \\
\midrule

\multirow{6}{*}{DAAC} 
 & MDI & 0.55 & 0.70 & X & 0.30 & 0.51 & 0.57 & 0.15 & 0.463 & \improvement{+3.9} \\
 & Depth Residuals + MDI & 0.54 & 0.65 & X & 0.28 & 0.51 & 0.53 & 0.15 & 0.443 & \improvement{+8.1} \\
 & OrC + MDI & 0.54 & 0.67 & X & 0.29 & 0.48 & 0.51 & \bg{\textbf{0.12}} & 0.435 &  \improvement{+9.8} \\
 & Depth Residuals + OrC + MDI & 0.54 & 0.65 & X & 0.28 & 0.55 & \textbf{0.50} & 0.15 & 0.445 & \improvement{+7.7} \\
 & Inverse DIFT & \sy{\underline{0.53}} & 0.72 & X & \bg{\textbf{0.25}} & 0.59 & 0.63 & 0.16 & 0.479 & \improvement{+0.6} \\
 & Metric DIFT & 0.57 & 0.68 & X & 0.29 & 0.71 & 0.59 & 0.14 & 0.496 & \improvement{-2.9} \\
\midrule

\multirow{4}{*}{VDA} 
 & MDI & 0.54 & 0.65 & X & 0.28 & 0.51 & 0.51 & 0.13 & 0.437 & \improvement{+9.3} \\
 & Depth Residuals + MDI & \bg{\textbf{0.52}} & 0.67 & \textbf{0.16} & \sy{\underline{0.27}} & \bg{\textbf{0.38}} & 0.51 & 0.15 & \sy{\underline{0.417}} & \improvement{+13.5} \\
 & OrC + MDI & 0.56 & \sy{\underline{0.60}} & \textbf{0.15} & 0.30 & 0.42 & 0.51 & 0.13 & 0.420 & \improvement{+12.9} \\
 & Depth Residuals + OrC + MDI & 0.55 & \textbf{\bg{0.57}} & \textbf{0.15} & 0.29 & 0.42 & 0.52 & 0.13 & \bg{\textbf{0.413}} & \improvement{+14.3} \\

\bottomrule
\end{tabular}
\end{table*}
\subsection{Experimental and  Implementation Details}
All experiments are conducted on an NVIDIA Jetson AGX Orin (64GB) running Jetpack 6.2 (L4T 36.4.4, CUDA 12.6). To ensure real-time performance, we export the ViT-S variants of the MDE models to FP16 TensorRT engines \cite{tensorrt}. We evaluate on the M3ED Spot sequences \cite{m3ed} to test quadruped motion, and the TartanGround Downtown sequences \cite{tartanground} again for quadruped motion in outdoor urban environments, reporting ATE RMSE against VINS-Mono \cite{vinsmono} after Sim(3) alignment.

DepthAnythingAC  achieves 12ms latency ($\sim$83 FPS), enabling synchronous front-end integration. In contrast, VideoDepthAnything incurs 44ms latency ($\sim$23 FPS) due to spatiotemporal overhead, necessitating asynchronous processing to preserve backend stability.

\subsection{Quantitative Analysis}
\label{sec:results}
The quantitative evaluation on the TartanGround (Table~\ref{tab:tartanground}) and M3ED (Table \ref{tab:m3ed}) datasets demonstrates that incorporating learned depth priors into the optimization backend can improve localization accuracy over the VINS-Mono baseline.

\noindent\textbf{Optimization vs. Injection:}
Our results highlight the efficacy of backend constraints over frontend injection. Integrating depth into the tracking frontend (DIFT) sometimes yielded negative results, while outperforming all methods in certain sequences (Table~\ref{tab:tartanground}). Conversely, backend optimization robustly handles noise as temporal consistency gating is applied. While Ordinal Priors excelled in structured environments, Affine and Ordinal priors combined proved most robust in real-world scenarios, achieving a \textbf{+14.3\% improvement} on the challenging M3ED dataset, as detailed in Table~\ref{tab:m3ed}.

\noindent\textbf{Impact of Temporal Stability (VDA vs. DAAC):}
VDA's proved superior for VIO integration. As shown in Table~\ref{tab:tartanground}, VDA with affine priors achieved the lowest ATE on the TartanGround Downtown with 0.478m, granting a \textbf{28.3\% improvement} over the baseline (0.676m). In contrast, the zero-shot DAAC model showed weaker gains (+23.4\% with best combination), confirming that inter-frame flicker in zero-shot models introduces noise that conflicts with optimization estimations.



\section{Conclusion \& Discussion}
\label{sec:conclusion}

In this work, we presented a real-time framework for enhancing monocular Visual-Inertial Odometry (VIO) using learned depth priors on edge devices. By integrating state-of-the-art MDE models into the VINS-Mono optimization backend, we enhanced the baseline performance while respecting the computational constraints of the NVIDIA Jetson AGX Orin. 

\noindent{\textbf{Key Insights.}} 
Our systematic evaluation on both synthetic and real-world datasets yields three critical insights for the design of future MDE-assisted VIO systems.

First, \textbf{temporal consistency is paramount}. While zero-shot models like DAAC offer low latency (12ms), their inter-frame flicker destabilizes optimization. In contrast, video-based models like VDA provide the temporal stability required for consistent trajectory estimation, achieving an error reduction of up to \textbf{28.3\%} on standard benchmarks.

Second, \textbf{backend optimization generally outperforms frontend injection}. While directly injecting depth into the visual tracking pipeline yielded the best performance in specific sequences, frequently degraded overall accuracy by corrupting photometric consistency. This failure mode, however, is largely attributable to the lack of temporal stability in real-time depth estimation. Consequently, incorporating depth as a weighted constraint in the backend factor graph currently provides a more robust mechanism for leveraging geometric priors,  while future advancements in temporally consistent low-latency depth inference may eventually unlock the efficacy of DIFT method.

Finally, \textbf{geometric priors can prevent catastrophic failure}. In challenging scenarios characterized by rapid motion or low texture (e.g., the M3ED 'Stairs' sequence), where the baseline VINS-Mono system diverged, our method successfully maintained a valid trajectory. This demonstrates that temporally consistent depth priors prove beneficial for sustaining navigation in environments where traditional approaches fail.

\section*{Acknowledgment}
This work is supported by TÜBİTAK 2224-A -- Grant Program for Participation in Scientific Meetings Abroad.

\newpage
\begingroup
    \small           
    \linespread{0.9}        
    \selectfont             
    \setlength{\itemsep}{0pt} 
    
    \bibliographystyle{IEEEbib}
    \bibliography{strings,refs}

@misc{droidslam,
      title={DROID-SLAM: Deep Visual SLAM for Monocular, Stereo, and RGB-D Cameras}, 
      author={Zachary Teed and Jia Deng},
      year={2022},
      eprint={2108.10869},
      archivePrefix={arXiv},
      primaryClass={cs.CV},
      url={https://arxiv.org/abs/2108.10869}, 
}

@misc{dpvo,
      title={Deep Patch Visual Odometry}, 
      author={Zachary Teed and Lahav Lipson and Jia Deng},
      year={2023},
      eprint={2208.04726},
      archivePrefix={arXiv},
      primaryClass={cs.CV},
      url={https://arxiv.org/abs/2208.04726}, 
}

@article{tightlycoupled,
title = {A tightly-coupled dense monocular Visual-Inertial Odometry system with lightweight depth estimation network},
journal = {Applied Soft Computing},
volume = {171},
pages = {112809},
year = {2025},
issn = {1568-4946},
doi = {https://doi.org/10.1016/j.asoc.2025.112809},
url = {https://www.sciencedirect.com/science/article/pii/S1568494625001206},
author = {Xin Wang and Zuoming Zhang and Luchen Li}
}

@ARTICLE{ordinal,
  author={Sun, Libo and Yin, Wei and Xie, Enze and Li, Zhengrong and Sun, Changming and Shen, Chunhua},
  journal={IEEE Transactions on Robotics}, 
  title={Improving Monocular Visual Odometry Using Learned Depth}, 
  year={2022},
  volume={38},
  number={5},
  pages={3173-3186},
  doi={10.1109/TRO.2022.3164834}}

@misc{dfvo,
      title={DF-VO: What Should Be Learnt for Visual Odometry?}, 
      author={Huangying Zhan and Chamara Saroj Weerasekera and Jia-Wang Bian and Ravi Garg and Ian Reid},
      year={2021},
      eprint={2103.00933},
      archivePrefix={arXiv},
      primaryClass={cs.CV},
      url={https://arxiv.org/abs/2103.00933}, 
}

@INPROCEEDINGS{mdvo,
  author={Li, Pengzhi and Tang, Chengshuai and Duan, Yifu and Li, Zhiheng},
  booktitle={2024 International Joint Conference on Neural Networks (IJCNN)}, 
  title={MD2VO: Enhancing Monocular Visual Odometry through Minimum Depth Difference}, 
  year={2024},
  volume={},
  number={},
  pages={1-8},
  doi={10.1109/IJCNN60899.2024.10649955}}

@inproceedings{dptvo,
   title={Dense Prediction Transformer for Scale Estimation in Monocular Visual Odometry},
   url={http://dx.doi.org/10.1109/LARS/SBR/WRE56824.2022.9995735},
   DOI={10.1109/lars/sbr/wre56824.2022.9995735},
   booktitle={2022 Latin American Robotics Symposium (LARS), 2022 Brazilian Symposium on Robotics (SBR), and 2022 Workshop on Robotics in Education (WRE)},
   publisher={IEEE},
   author={Francani, Andre O. and Maximo, Marcos R. O. A.},
   year={2022},
   month=oct, pages={1–6} }

@misc{triangulation,
      title={Deep Visual Odometry and Pose Reconstruction through Single
Image Depth Map and Triangulation}, 
      author={Stefano Silvestrini},
      year={2024},
      url={https://re.public.polimi.it/retrieve/c8e1ee02-236b-46c2-90c8-eaf1ae20a2de/SILVS03-24.pdf}, 
}

@misc{onlinetrain,
      title={An Online Adaptation Method for Robust Depth Estimation and Visual Odometry in the Open World}, 
      author={Xingwu Ji and Haochen Niu and Dexin Duan and Rendong Ying and Fei Wen and Peilin Liu},
      year={2025},
      eprint={2504.11698},
      archivePrefix={arXiv},
      primaryClass={cs.RO},
      url={https://arxiv.org/abs/2504.11698}, 
}

@misc{pseudorgb,
      title={Pseudo RGB-D for Self-Improving Monocular SLAM and Depth Prediction}, 
      author={Lokender Tiwari and Pan Ji and Quoc-Huy Tran and Bingbing Zhuang and Saket Anand and Manmohan Chandraker},
      year={2020},
      eprint={2004.10681},
      archivePrefix={arXiv},
      primaryClass={cs.CV},
      url={https://arxiv.org/abs/2004.10681}, 
}

@INPROCEEDINGS{carlos,
  author={Campos, Carlos and Tardós, Juan D.},
  booktitle={2022 IEEE/RSJ International Conference on Intelligent Robots and Systems (IROS)}, 
  title={Scale-aware direct monocular odometry}, 
  year={2022},
  volume={},
  number={},
  pages={1360-1366},
  doi={10.1109/IROS47612.2022.9982173}}

@misc{dav2,
      title={Depth Anything V2}, 
      author={Lihe Yang and Bingyi Kang and Zilong Huang and Zhen Zhao and Xiaogang Xu and Jiashi Feng and Hengshuang Zhao},
      year={2024},
      eprint={2406.09414},
      archivePrefix={arXiv},
      primaryClass={cs.CV},
      url={https://arxiv.org/abs/2406.09414}, 
}

@misc{dav3,
      title={Depth Anything 3: Recovering the Visual Space from Any Views}, 
      author={Haotong Lin and Sili Chen and Junhao Liew and Donny Y. Chen and Zhenyu Li and Guang Shi and Jiashi Feng and Bingyi Kang},
      year={2025},
      eprint={2511.10647},
      archivePrefix={arXiv},
      primaryClass={cs.CV},
      url={https://arxiv.org/abs/2511.10647}, 
}

@misc{depthpro,
      title={Depth Pro: Sharp Monocular Metric Depth in Less Than a Second}, 
      author={Aleksei Bochkovskii and Amaël Delaunoy and Hugo Germain and Marcel Santos and Yichao Zhou and Stephan R. Richter and Vladlen Koltun},
      year={2025},
      eprint={2410.02073},
      archivePrefix={arXiv},
      primaryClass={cs.CV},
      url={https://arxiv.org/abs/2410.02073}, 
}

@article{metric3dv2,
   title={Metric3D v2: A Versatile Monocular Geometric Foundation Model for Zero-Shot Metric Depth and Surface Normal Estimation},
   volume={46},
   ISSN={1939-3539},
   url={http://dx.doi.org/10.1109/TPAMI.2024.3444912},
   DOI={10.1109/tpami.2024.3444912},
   number={12},
   journal={IEEE Transactions on Pattern Analysis and Machine Intelligence},
   publisher={Institute of Electrical and Electronics Engineers (IEEE)},
   author={Hu, Mu and Yin, Wei and Zhang, Chi and Cai, Zhipeng and Long, Xiaoxiao and Chen, Hao and Wang, Kaixuan and Yu, Gang and Shen, Chunhua and Shen, Shaojie},
   year={2024},
   month=dec, pages={10579–10596} }

@misc{unidepth,
      title={UniDepth: Universal Monocular Metric Depth Estimation}, 
      author={Luigi Piccinelli and Yung-Hsu Yang and Christos Sakaridis and Mattia Segu and Siyuan Li and Luc Van Gool and Fisher Yu},
      year={2024},
      eprint={2403.18913},
      archivePrefix={arXiv},
      primaryClass={cs.CV},
      url={https://arxiv.org/abs/2403.18913}, 
}

@article{unidepthv2,
   title={UniDepthV2: Universal Monocular Metric Depth Estimation Made Simpler},
   ISSN={1939-3539},
   url={http://dx.doi.org/10.1109/TPAMI.2025.3628473},
   DOI={10.1109/tpami.2025.3628473},
   journal={IEEE Transactions on Pattern Analysis and Machine Intelligence},
   publisher={Institute of Electrical and Electronics Engineers (IEEE)},
   author={Piccinelli, Luigi and Sakaridis, Christos and Yang, Yung-Hsu and Segu, Mattia and Li, Siyuan and Abbeloos, Wim and Van Gool, Luc},
   year={2025},
   pages={1–14} }

@misc{daac,
      title={Depth Anything at Any Condition}, 
      author={Boyuan Sun and Modi Jin and Bowen Yin and Qibin Hou},
      year={2025},
      eprint={2507.01634},
      archivePrefix={arXiv},
      primaryClass={cs.CV},
      url={https://arxiv.org/abs/2507.01634}, 
}

@misc{moge,
      title={MoGe: Unlocking Accurate Monocular Geometry Estimation for Open-Domain Images with Optimal Training Supervision}, 
      author={Ruicheng Wang and Sicheng Xu and Cassie Dai and Jianfeng Xiang and Yu Deng and Xin Tong and Jiaolong Yang},
      year={2025},
      eprint={2410.19115},
      archivePrefix={arXiv},
      primaryClass={cs.CV},
      url={https://arxiv.org/abs/2410.19115}, 
}

@misc{vda,
      title={Video Depth Anything: Consistent Depth Estimation for Super-Long Videos}, 
      author={Sili Chen and Hengkai Guo and Shengnan Zhu and Feihu Zhang and Zilong Huang and Jiashi Feng and Bingyi Kang},
      year={2025},
      eprint={2501.12375},
      archivePrefix={arXiv},
      primaryClass={cs.CV},
      url={https://arxiv.org/abs/2501.12375}, 
}

@misc{eigen,
      title={An Introduction to Convolutional Neural Networks}, 
      author={Keiron O'Shea and Ryan Nash},
      year={2015},
      eprint={1511.08458},
      archivePrefix={arXiv},
      primaryClass={cs.NE},
      url={https://arxiv.org/abs/1511.08458}, 
}

@misc{metric3d,
      title={Metric3D: Towards Zero-shot Metric 3D Prediction from A Single Image}, 
      author={Wei Yin and Chi Zhang and Hao Chen and Zhipeng Cai and Gang Yu and Kaixuan Wang and Xiaozhi Chen and Chunhua Shen},
      year={2023},
      eprint={2307.10984},
      archivePrefix={arXiv},
      primaryClass={cs.CV},
      url={https://arxiv.org/abs/2307.10984}, 
}

@article{vinsmono,
   title={VINS-Mono: A Robust and Versatile Monocular Visual-Inertial State Estimator},
   volume={34},
   ISSN={1941-0468},
   url={http://dx.doi.org/10.1109/TRO.2018.2853729},
   DOI={10.1109/tro.2018.2853729},
   number={4},
   journal={IEEE Transactions on Robotics},
   publisher={Institute of Electrical and Electronics Engineers (IEEE)},
   author={Qin, Tong and Li, Peiliang and Shen, Shaojie},
   year={2018},
   month=aug, pages={1004–1020} }

@article{orbslam,
   title={ORB-SLAM3: An Accurate Open-Source Library for Visual, Visual–Inertial, and Multimap SLAM},
   volume={37},
   ISSN={1941-0468},
   url={http://dx.doi.org/10.1109/TRO.2021.3075644},
   DOI={10.1109/tro.2021.3075644},
   number={6},
   journal={IEEE Transactions on Robotics},
   publisher={Institute of Electrical and Electronics Engineers (IEEE)},
   author={Campos, Carlos and Elvira, Richard and Rodriguez, Juan J. Gomez and M. Montiel, Jose M. and D. Tardos, Juan},
   year={2021},
   month=dec, pages={1874–1890} }

@INPROCEEDINGS{m3ed,
  author={Chaney, Kenneth and Cladera, Fernando and Wang, Ziyun and Bisulco, Anthony and Hsieh, M. Ani and Korpela, Christopher and Kumar, Vijay and Taylor, Camillo J. and Daniilidis, Kostas},
  booktitle={2023 IEEE/CVF Conference on Computer Vision and Pattern Recognition Workshops (CVPRW)}, 
  title={M3ED: Multi-Robot, Multi-Sensor, Multi-Environment Event Dataset}, 
  year={2023},
  volume={},
  number={},
  pages={4016-4023},
  doi={10.1109/CVPRW59228.2023.00419}}

@misc{tartanground,
      title={TartanGround: A Large-Scale Dataset for Ground Robot Perception and Navigation}, 
      author={Manthan Patel and Fan Yang and Yuheng Qiu and Cesar Cadena and Sebastian Scherer and Marco Hutter and Wenshan Wang},
      year={2025},
      eprint={2505.10696},
      archivePrefix={arXiv},
      primaryClass={cs.RO},
      url={https://arxiv.org/abs/2505.10696}, 
}

@misc{vit,
      title={An Image is Worth 16x16 Words: Transformers for Image Recognition at Scale}, 
      author={Alexey Dosovitskiy and Lucas Beyer and Alexander Kolesnikov and Dirk Weissenborn and Xiaohua Zhai and Thomas Unterthiner and Mostafa Dehghani and Matthias Minderer and Georg Heigold and Sylvain Gelly and Jakob Uszkoreit and Neil Houlsby},
      year={2021},
      eprint={2010.11929},
      archivePrefix={arXiv},
      primaryClass={cs.CV},
      url={https://arxiv.org/abs/2010.11929}, 
}

@misc{dinov2,
      title={DINOv2: Learning Robust Visual Features without Supervision}, 
      author={Maxime Oquab and others},
      year={2024},
      eprint={2304.07193},
      archivePrefix={arXiv},
      primaryClass={cs.CV},
      url={https://arxiv.org/abs/2304.07193}, 
}

@misc{dinov3,
      title={DINOv3}, 
      author={Oriane Siméoni and others},
      year={2025},
      eprint={2508.10104},
      archivePrefix={arXiv},
      primaryClass={cs.CV},
      url={https://arxiv.org/abs/2508.10104}, 
}

@manual{tensorrt,
  author = {{NVIDIA Corporation}},
  title = {NVIDIA TensorRT: Programmable Inference Accelerator},
  year = {2024},
  note = {Version 10.3.0},
  url = {https://developer.nvidia.com/tensorrt}
}

@book{slam-handbook,
      title      = {{SLAM Handbook.} From Localization and Mapping to Spatial Intelligence},
      editor     = {Luca Carlone and Ayoung Kim and Timothy Barfoot and Daniel Cremers and Frank Dellaert},
      publisher  = {Cambridge University Press},
      year       = {2025}
    }

@misc{peng2022rwtslamrobustvisualslam,
      title={RWT-SLAM: Robust Visual SLAM for Highly Weak-textured Environments}, 
      author={Qihao Peng and Zhiyu Xiang and YuanGang Fan and Tengqi Zhao and Xijun Zhao},
      year={2022},
      eprint={2207.03539},
      archivePrefix={arXiv},
      primaryClass={cs.CV},
      url={https://arxiv.org/abs/2207.03539}, 
}
\endgroup

\end{document}